# Broken Rail Detection With Texture Image Processing Using Two-Dimensional Gray Level Co-occurrence Matrix

**Mohsen Ebrahimi[1]**

*1 Ph.D. student of control and system engineering, Department of Electric and Computer, Malek Ashtar University of Technology, Tehran, Iran, Mohsen_ebrahimi@mut.ac.ir*
*\* Corresponding author email: Mohsen_ebrahimi@mut.ac.ir*

## Abstract

Application of electronic railway systems as well as the implication of Automatic Train Control (ATC) System has increased the safety of rail transportation. However, one of the most important causes of accidents on the railway is rail damage and breakage. In this paper, we have proposed a method that the rail region is first recognized from the observation area, then by investigating the image texture processing data, the types of rail defects including cracks, wear, peeling, disintegration, and breakage are detected. In order to reduce the computational cost, the image is changed from the RGB color spectrum to the gray spectrum. Image texture processing data is obtained by the two-dimensional Gray Levels Co-occurrence Matrix (GLCM) at different angles; this data demonstrates second-order features of the images. Large data of features has a negative effect on the overall accuracy of the classifiers. To tackle this issue and acquire faster response, Principal Component Analysis (PCA) algorithm is used, before entering the band into the classifier. Then the features extracted from the images are compared by three different classifiers including Support Vector Machine (SVM), Random Forest (RF), and K-Nearest Neighbor (KNN) classification. The results obtained from this method indicate that the Random Forest classifier has better performance (accuracy 97%, precision 96%, and recall 96%) than other classifiers.

# 1- Introduction

The requirements for rail operations' safety are becoming more and more significant as train speed and carrying capacity both increase. A variety of defects in varying degrees gradually appear on the track surface as a result of the influence of temperature, humidity, load, and other factors; if the defects are not corrected in a timely manner, the degree of defects will worsen and significantly raise the risk of train operation. The traditional method of inspecting the surface of railroad tracks relies primarily on visual inspection, but this method is risky for the inspectors and has other drawbacks such as implementation errors, a long turnaround time, high costs, and exposure to the harsh environment [1]. The development of automatic rail surface defect detection is thus crucial for both practical use and research.

Although the causes of the failure of the rail surface can vary, they can all be traced back to the rail's contact with the wheel. The existence of surface defects on the rail does not imply that the track is dangerous. However, if it is not resolved quickly, it may result in more serious issues like rail breakage and, in the long run, rail vehicle derailment. As a result, it is important to regularly check for rail surface defects, which is also one of the main considerations when planning rail maintenance. The installation of various types of sensors on an inspection wagon or a rail vehicle is the typical foundation for the automated track inspection systems that have been developed over time. The difficulties of precisely detecting surface defects have not yet been overcome by these systems, which are primarily based on laser, sound emission, and ultrasonic wave technologies [2,3]. LIDAR [4] and ground penetrating radar [5]-based techniques are useful for spotting or keeping an eye on relatively large targets, like railroad pylons and ballast, but they cannot be used for small targets, like rail surfaces.

The use of artificial intelligence technology in the railway industry has become a research focus in recent years [6–8] due to the rapid development of machine vision and deep learning. Automatically identifying rail surface flaws in the rail image is achieved by the computer vision detection method using machine vision technology [9,10]. It is an effective replacement for the conventional visual method and conserves both human and material resources [11,12].

Some target detection algorithms have also been applied in the field of fault detection, where deep learning and convolutional neural networks have been used in the field of images [13,14]. Zheng and colleagues. [15] proposed a deep learning algorithm with compression and expansion mechanisms for fault detection on copper boards with quick detection speed. The pre-trained network VGG19 was employed in [16] to find flaws in Li-ion battery electrodes. Hao and co. [17] detected steel surface defects and accomplished localization and classification of steel surface defects using a deformable convolutional augmented backbone network and feature pyramid.

To improve the contrast of rail images for feature extraction, a local normalization algorithm was suggested in [18]. These techniques have been shown to be particularly effective at identifying rail flaws. However, one of their main drawbacks is that a number of flaws, such as linear flaws, cracks, and small flaws, are challenging to detect, and the results of the detection are typically inaccurate and imprecise. Wang and others. [19] suggested using color features in conjunction with principal component analysis (PCA) to detect rail surface defects. To identify rail defects, [20] proposed a sequenced combination of the gray balance model of phase spectrum and Otsu's threshold segmentation method. Deep convolutional neural networks (CNNs) for automatic feature extraction have been introduced in recent years, with good accuracy and productivity thanks to the rapid development of these networks. With the help of these robust capabilities, deep CNNs have been successfully used in several studies [21] to classify rail surface defects. These methods have the drawback that they are unable to perform quick inference in complex models.

Gray-level co-occurrence matrices (GLCMs) are one of the most well-known and common image texture feature extraction methods. The GLCM method is one of the second-order feature extraction methods because it is based on the convolutional relationship between adjacent pixels. GLCM characterizes the texture of an image by repeatedly calculating different values of adjacent pixel pairs, at different angles, and then extracting texture features such as energy, entropy, contrast, homogeneity, etc. Reference [22,23] has used co-occurrence matrices to detect surface defects. Most of the use of this method is in the field of identifying defects in fabric [24,25]. Despite the many advantages of GLCM, it has several shortcomings. For example, in this method, there is no accepted optimization for the displacement vector and the extracted GLCM features have statistical and computational dependencies.

In this article, a modified dataset containing 1700 images including three classes, healthy, defective, and junction, is used. which are almost equally divided between these three classes. In this data set, the rail area is first separated from the whole image and pre-processed using noise removal filters, contrast improvement, and histogram matching to extract its features by the GLCM method. In the GLCM method, for all data sets, the features are obtained with four offsets with different angles of 0, 45, 90, and 135 degrees, and their average is taken. One of the innovations of this article is the optimization of features using the PCA method, which reduces the volume of calculations and increases the response speed. The optimized features are divided by different classifiers and the results are displayed in an example.

In the continuation of the process of the article, in the first part, explanations regarding the pre-processing performed on the images received from the dataset are described. Mask production to separate the rail track from the image and also the challenges that exist in this direction will be explained. Then, after the initial processing of the data set, it entered the feature extraction section from the images, where explanations are given on how to use the GLCM method and its offsets. Considering that optimization methods have not been applied to GLCM in the previous methods, in the fourth section, the PCA optimization method is discussed and how to reduce features in an optimal way is described. At the end, the correctness of the method is proven by simulating a real example, and the results obtained from this algorithm in the simulation are stated in the conclusion section.

## 2- Method and process
In this section, the algorithm is used and its different parts will be explained.

### 2-1- Preprocessing

After receiving the images from the dataset, preprocessing is one of the important and key parts of any pattern processing system. It is important because the failure to apply appropriate approaches at this stage causes the spread of errors in the entire system and ultimately affects the efficiency of the entire system. The steps to be taken until reaching the result are shown in the flowchart of Figure 1.

Using thresholding methods and according to the input data set, to separate the rail range from the image, various filters such as the Atsu filter and segmentation functions are applied to the data set so that the required range is separated from the entire image and additional processing is not performed. For this purpose, after producing the required mask, additional areas are removed from the image.

In the pre-processing stage, the image noise should be taken and the light of the image should be improved by means of different filters, depending on the desired need. Before entering the filter section, all the images are converted to the same dimensions so that all the processing can be done according to the desired and suitable dimensions. Equalizing the dimensions of the image also helps to increase the processing speed, which is because all the calculations happen in the format specified for it. The combined use of Gaussian, Laplacian, and Madin filters according to the type of data set can help in this section.

An improved variant of the histogram equalizer (HE) is the adaptive histogram equalizer (AHE). AHE performs histogram equalization in small areas in the image and increases the contrast of each area separately, which is used in this article.

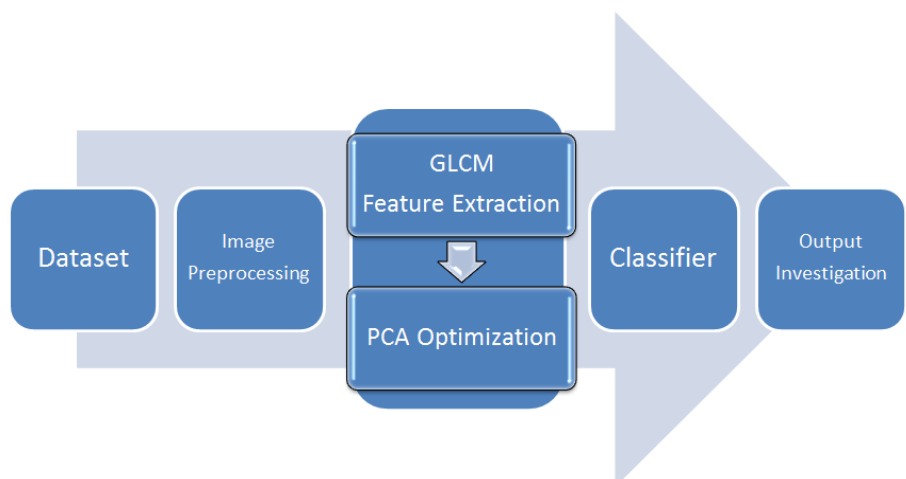

**Figure 1**. Flowchart of algorithm steps

### 2-2- Feature extraction
Image texture analysis by statistical indices extracted from the histogram, although it provides useful information about the spatial distribution of the gray levels of pixels in the image, it does not provide any information about the relative spatial relationship of the values of the gray levels of the pixels. The reason for this is the insensitivity of the histogram values to the displacement of the image pixels. The second-order statistical indices consider the relationship between the values of the gray levels of two pixels at a certain distance and direction from each other. The gray level simultaneity matrix is a square matrix that can describe the texture characteristics by calculating the gray spatial correlation of the image [26].

The GLCM texture, which represents the two-dimensional statistical properties of the image texture, takes into account the spatial relationship between two pixels in a particular orientation angle and distance [27]. An appropriate gray-level concurrency matrix for the offset value and direction in GLCM should be provided. In four directions of 0°, 45°, 90°, and 135° when the offset value is 1, a gray-level co-occurrence matrix is present [26]. The number of occurrences of the two gray values represented by the rows and columns in the image occurring together at the specified offset value and direction is the value of the gray-level concurrency matrix element. The spatial correlation features of images are reflected by 14 GLCM features. Energy, entropy, contrast, and homogeneity are four of these characteristics that are frequently connected to texture [28].

Below are the relationships between these features separately:

$$Contrast = \sum_{i}^{M}\sum_{j}^{N}(i-j)^2 p[i,j] \qquad (1)$$

$$Correlation = \frac{\sum_{i}^{M}\sum_{j}^{N}(i-\mu)(j-\mu)p[i,j]}{\sigma^2} \qquad (2)$$

$$Energy = \sum_{i}^{M}\sum_{j}^{N}(p[i,j])^2 \qquad (3)$$

$$Homogenety = \sum_{i}^{M}\sum_{j}^{N}\frac{p[i,j]}{1+|i-j|} \qquad (4)$$

Where M, N are the dimensions of the image $p[i,j]$ for $i=1,2,3,…, M$, and $j=1,2,…,N$. $\mu$ and $\sigma$ are the mean and the variance of the image $p[i,j]$.

### 2-3- Principal components analysis

The main elements of a set of points in a real coordinate space are a series of unit vectors, where the i-th vector is the line extension that best fits the data and is orthogonal to the first i-1 vector. The best line in this situation is the one that minimizes the mean square of the angles at which the points are perpendicular to the line. The different dimensions of the data are not linearly correlated because these vectors form an orthogonal basis. The process of calculating the principal components and transforming the database using them, sometimes using only the first few principal components and ignoring the rest, is known as principal component analysis. Operations for feature reduction are one of PCA's primary applications.

In fact, PCA extracts those features that provide more value to us. The principal component analysis is actually a statistical technique to reduce the dimensions of a data set. This is accomplished by linearly transforming the data into a new coordinate system, where (most) changes in the data can be described with fewer dimensions than the original data. The main idea of this method is to find orthogonal components that express the largest amount of variance possible. In practice, we usually want to represent the data by extracting a few important components in smaller dimensions so that most of the structures in the data are preserved.

The optimal dimensions for the PCA method are obtained by trial and error, and in this paper, by doing this, we reached the numbers 50 and 20 for this dataset, both of which provided suitable answers. As a result, to speed up the response, the number 20 was applied for PCA dimensions in the simulation.

## 2-4- Classifiers

For data classification, in this article, three classifiers of SVM[1], RF[2], and KNN[3] are used, and also the results of a 15-layer neural network are presented using the nprtool neural network toolbox. In order to increase the response, the number of hidden layers of the neural network can be increased, although this works until the neural network does not reach saturation, that is, it reaches a point where increasing the number of layers does not have an effect on improving the response. The classifiers categorize the data based on accuracy, sensitivity, specificity, precision, F-mean, and g-mean, whose equation is given in relations 5, 6, 7, 8, 9, and 10, respectively.

$$Accuracy = \frac{TP + TN}{(TP + FP) + (TN + FN)} \quad (5)$$

$$Sensitivity = \frac{TP}{(TP + FN)} \quad (6)$$

$$Specificity = \frac{TP}{(FP + TN)} \quad (7)$$

$$Precision = \frac{TP}{(TP + FP)} \quad (8)$$

$$F - mean = \frac{2TP}{(2TP + FP + FN)} \quad (9)$$

$$g - mean = \sqrt{TPR * TNR} \quad (10)$$

In these relations, T and F are symbols of true and false, respectively, and P and N are symbols of positive and negative. And:

$$TPR = \frac{TP}{TP + FN} \quad (11)$$

$$TNR = \frac{TN}{TN + FP} \quad (12)$$

## 3- simulation

The input data set contains three classes with healthy, defective, and junction labels, an example of these three classes is shown in Figure 2, At first, the rail part is separated from the image space of the data set using thresholding and masking methods. Due to the fact that the camera installed on the train to collect the data set and the train moving on the straight track, the margins created in the images are similar and it is possible to separate the rail track with the same mask. In Figure 3 the mask created and applied on a sample of the data set can be seen. Then the process of pre-processing the image, equalizing the dimensions, and correcting the spectrum is done using different filters on the dataset.

---

[1] support vector machine

[2] random forest

[3] K nearest neighbor

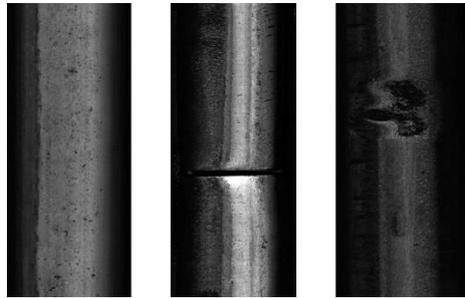

**Figure 2.** An example of an image of a data class, the left is healthy, the middle is joint, and the right is defects

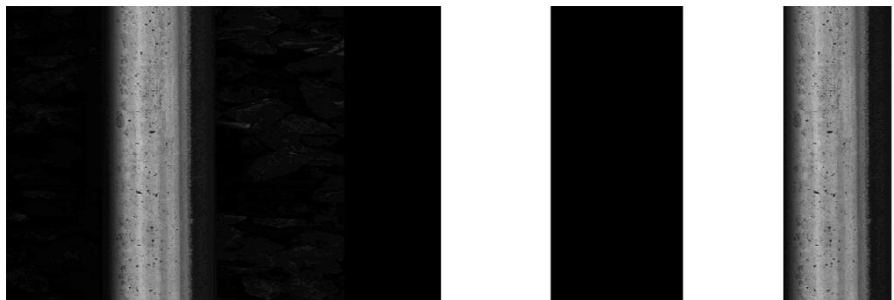

**Figure 3**. Performing a mask to separate the rail range from the image

After the pre-processing, the images are ready to enter the feature extraction section. Figure 4 shows an example of the data set images after pre-processing and the changes that occur on the histogram of the image.

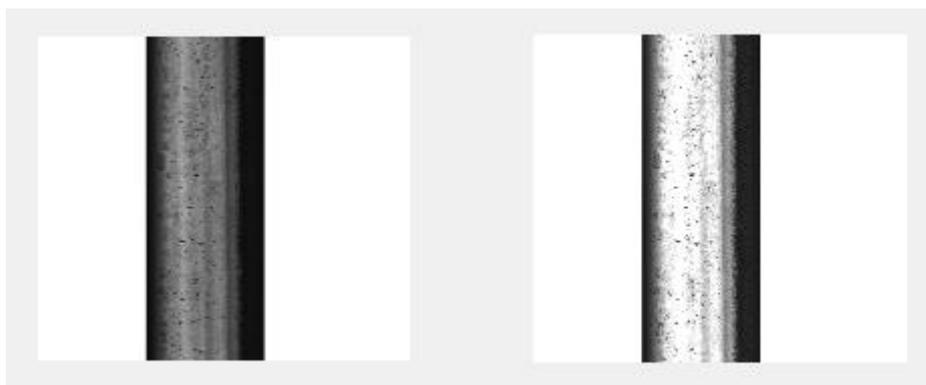

Figure 4: Effect of preprocessing on images

It can be seen in Figure 4 that the surface of the rail is displayed more clearly after pre-processing. By extracting the features and optimizing them, it is entered into the classifiers section and the results are obtained as shown in Figure 5

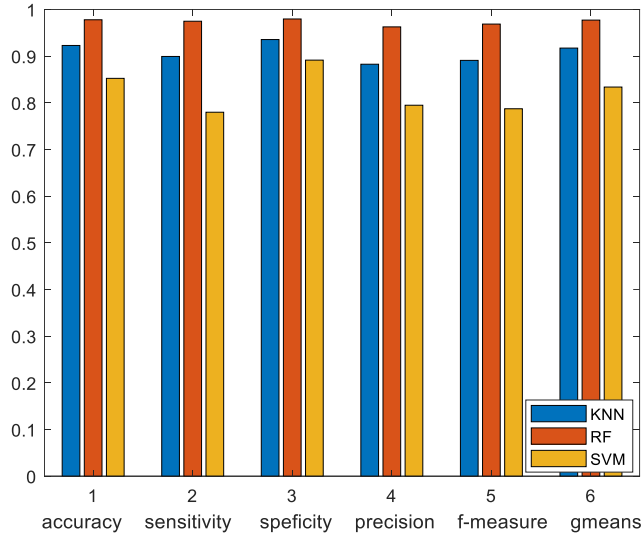

**Figure 5**. The bar chart of the result obtained from the simulation of the algorithm

The exact numbers of the diagram in Figure 5 are listed separately in Table 1.

Table 1. The results of the classifiers

|  | KNN | RF | SVM |
|---|---|---|---|
| **accuracy** | 0.92291 | 0.97977 | 0.85242 |
| **sensitivity** | 0.89937 | 0.96855 | 0.77987 |
| **specificity** | 0.93559 | 0.97966 | 0.89153 |
| **precision** | 0.88272 | 0.9625 | 0.79487 |
| **F-mean** | 0.89097 | 0.96552 | 0.7873 |
| **g-mean** | 0.9173 | 0.97409 | 0.83383 |

The simulation results show that the performance of the RF classifier is better than other classifiers.

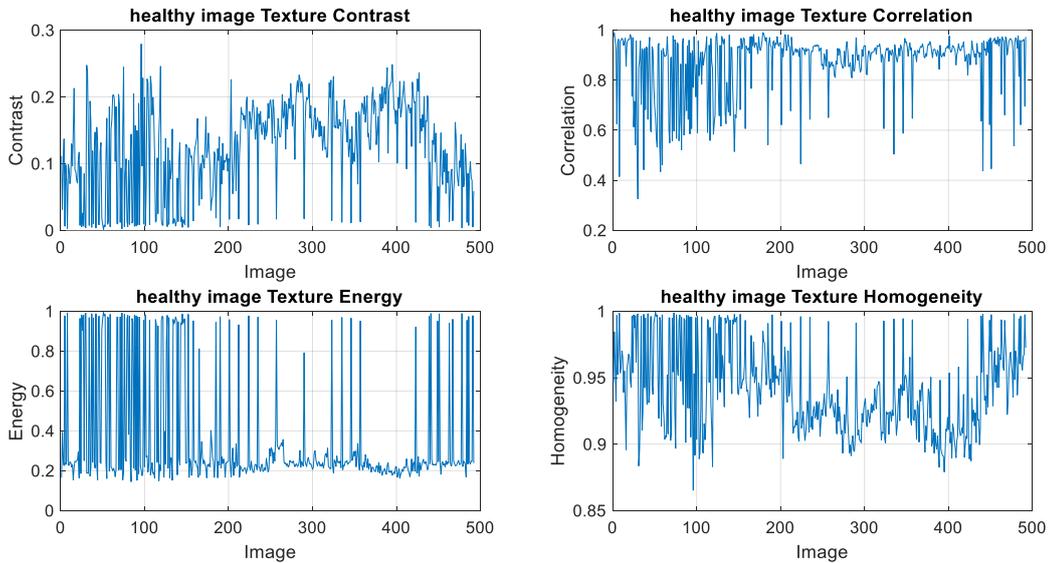

**Figure 6**. The plot of statistical properties of GLCM for healthy class

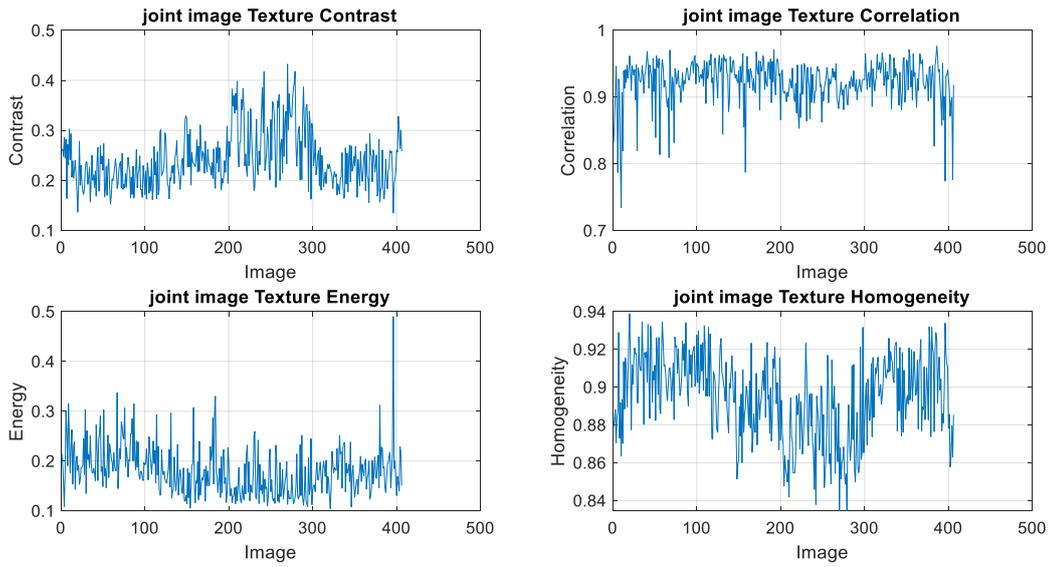

**Figure 7**. The plot of GLCM statistical properties for a joint class

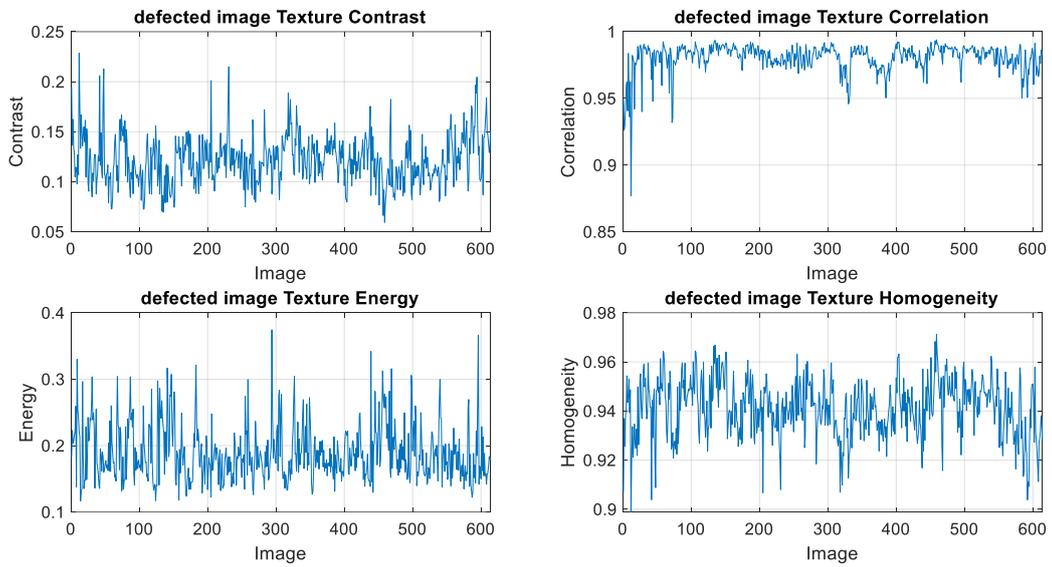

**Figure 8**. The plot of statistical properties of GLCM for defective class

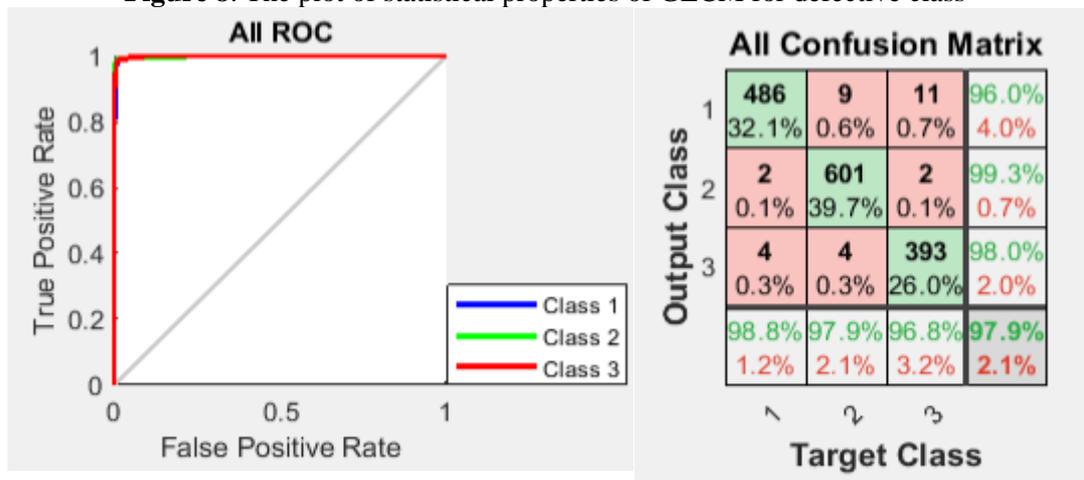

**Figure 9**. The result of the 15-layer neural network classifier

٨

As seen in Figure 9, the 15-layer neural network also obtains an accurate response that is very close to the response of the RF classifier. Since the accuracy of performance is very important in this problem, the best classifier is the classifier that gives higher accuracy results.

## 4- Conclusion

In this article, the classification of rail images based on three different classes, healthy, connected, and defective, was investigated for the selected dataset. In the pre-processing section, after thresholding for the image and separating the Rayleigh part from the data set images, from three combined Gaussian, Median, and Laplacian filters, where the image features using the gray co-occurrence matrix method (GLCM) along with the statistical features of the image histogram Extracted, used. After PCA optimization, the features were provided to different classifiers. The use of PCA makes the answer faster, therefore the program execution time is reduced. The classifiers used in this thesis include KNN, neural networks, SVM, and RF. The results of the simulations show that the RF classifier had a better performance.
To continue and improve this algorithm, it is suggested to use other combined filters in the pre-processing stage according to the selected dataset, to use optimization algorithms such as PSO or genetics to optimize data instead of PCA, and to check the response and innovation in floor design. New agents should be focused to achieve a better response.